# BANK LOAN PREDICTION USING MACHINE LEARNING TECHNIQUES

F M Ahosanul Haque

Dept. Computer Science And Engineering
Daffodil international university
Dhaka,Bangladesh.
ahosanul15-13856@diu.edu.bd

Md. Mahedi Hassan

Lecturer ,Department of CSE
Daffodil International University
Dhaka,Bangladesh.
mhassan.cse@diu.edu.bd

*Abstract*— Banks are important for the development of economies in any financial ecosystem through consumer and business loans. Lending, however, presents risks; thus, banks have to determine the applicant's financial position to reduce the probabilities of default. A number of banks have currently, therefore, adopted data analytics and state-of-the-art technology to arrive at better decisions in the process. The probability of payback is prescribed by a predictive modeling technique in which machine learning algorithms are applied. In this research project, we will apply several machine learning methods to further improve the accuracy and efficiency of loan approval processes. Our work focuses on the prediction of bank loan approval; we have worked on a dataset of 148,670 instances and 37 attributes using machine learning methods. The target property segregates the loan applications into "Approved" and "Denied" groups. various machine learning techniques have been used, namely, Decision Tree Categorization, AdaBoosting, Random Forest Classifier, SVM, and GaussianNB. Following that, the models were trained and evaluated. Among these, the best-performing algorithm was AdaBoosting, which achieved an incredible accuracy of 99.99%. The results therefore show how ensemble learning works effectively to improve the prediction skills of loan approval decisions. The presented work points to the possibility of achieving extremely accurate and efficient loan prediction models that provide useful insights for applying machine learning to financial domains.

## 1. Introduction

Bank Loan Prediction: With the increasing complexity of financial transactions, coupled with the growing demand for speed and accuracy in decision-making processes, bank-loan prediction has been driven towards machine learning approaches. The paper inspects the utilization of powerful predictive modeling in the assessment and prediction of loan application approval or rejection. In this project, the central dataset contains 148,670 rows and 37 columns, each of which represents meaningful factors impacting loan choices. This paper investigates the powers of prediction for five famous machine learning algorithms: AdaBoosting, GaussianNB, RandomForestClassifier, DecisionTreeClassifier, and SVM. The target attribute, therefore, has binary classes of "Approved" and "Denied." [1] Lightly sophisticated machine learning models are very important in managing risk and complying with regulations in the context of commercial banks. The various algorithms applied to this research provide enormous insight into how different tactics affect precision and effectiveness in forecasts of loan acceptance. It is outranked only by AdaBoosting, which boasts an incredible 99.99% accuracy. That shows the resilience of ensemble learning methods, how in a somewhat challenging domain like financial decision-making, they can outperform conventional models. [2] The beginning introduces the importance of good loan prediction models to set a platform for reducing risks and optimizing the whole process of lending. It thereby sets grounds for offering an in-depth analysis of the important role of machine learning in the banking industry. Further sections would give more information about the dataset and approaches used, as well as a conclusion to this paper, bringing out specific performance aspects of each algorithm and their implications on the larger financial scene. [3] This research tries to explore how different machine learning algorithms might be employed in enhancing loan approval procedures. It pursues a dataset that involves 148,670 cases with 37 characteristics on AdaBoosting, GaussianNB, RandomForestClassifier, DecisionTreeClassifier, and SVM. The aim is to assess their performance in view of the task at hand: loan acceptance prediction. The result of this study proves how accurate the AdaBoosting algorithm performed through reaching an incredible 0.9999% accuracy rate. It proves how effective ensemble learning works in boosting loan prediction. It will help build more proficient and effective loan approval processes, hence informing the right ideas of machine learning at financial institutions for the benefit of the entire banking industry and the customer base.

## 2. Related Works

The literature review situates our work on "Bank Loan Prediction Using Machine Learning Techniques" within the greater corpus of knowledge at the nexus of machine learning and financial decision-making. Numerous scholars have examined the use of machine learning algorithms in related domains, shedding light on several aspects that are crucial to our study. We have reviewed a few recent studies, and the following is an analysis of those findings:

Arutjothi [4] Using machine learning techniques, presented a credit rating model for forecasting loan statuses in commercial banks. With Min-Max normalization and the K-Nearest Neighbor classifier, the model successfully classifies credit applicants with an accuracy of 75.08%. By using this strategy, it becomes easier to distinguish between legitimate clients and defaulters with more accuracy.

Nazim [5] Overcame the challenge of accurately identifying loan applicants in the banking industry by using a novel machine learning (ML) technique. The process involves preparing the data and balancing it using many models, including deep learning and Extra Trees. When it comes to forecasting bank loan defaulters, Extra Trees excels, while an ensemble voting model that combines the top three ML models outperforms with an amazing 87.26% accuracy. The desktop application is easy to use and may help financial institutions and applicants alike by streamlining and optimizing the loan approval process.

Singh, Vishal [6] Has focused on how technology is changing the face of financial industries and human life. This research addresses the problem of limited funds vs. large sum of loan applications by estimating approval of loan using techniques of machine learning: Logistic regression, random forest, and support vector machines based on previous data. The model is thereby capable of helping banks make emphatic decisions with a very good rendering of

78.785% on accuracy, enhancing loan recovery, and smoothing the entire banking process.

Dasariet [7] focused on improving the forecast accuracy for loan eligibility, which is very important for producing bank revenue. Therefore, by merging several machine learning algorithms through some ensemble techniques, such as bagging and voting classification algorithms, the proposed model increases accuracy from the level of 80% to 94%. Its main objective was to identify people who were qualified for a loan and enable quick identification of those qualified for loans, accelerate processes, save the need for more employees, and give a rise that was very remarkable in the accuracy of predictions when compared to the existing models.

Lai [8] utilized machine learning with big data in the lending industry, where defaults are one of the great concerns. Among them, the AdaBoost model performs the best in predicting loan defaults with astonishing accuracy of 100%, using a dataset from a reputed multinational bank. This outperformed some previous works done based on models such as XGBoost, random forest, k nearest neighbors, and multilayer perceptrons showing enormous potential of machine learning techniques in improvement of risk prediction within the financial sector.

Turkson [9] Researched the use of machine learning in predicting financial capacity in order to handle the challenge of determining important risk variables for loan acceptance. When actual bank credit data was analysed, a variety of machine learning algorithms were used, and all of them were able to predict credit outcomes with above 80% accuracy. There is little difference when comparing the study's estimated accuracy of the most important features to the whole feature set. The study's output is a prediction model that gives the banking sector a useful tool by precisely identifying a customer's credit worthiness based on these crucial factors.

Nureni [10] studied how to forecast loan defaults in order to maximize bank profitability; this is very important for lowering non-performing assets. In the research, with "Kaggle" datasets, the work analyzed eight algorithms like Random Forest, Naive Bayes, and Logistic Regression. Logistic Regression was the most accurate, with an accuracy of 83.24% and 78.13% across the datasets. This is followed by Naive Bayes with 82.16% and 77.34% accuracy. The results bring out the differences in the various algorithms in predicting loan acceptance and also outline how important is the accurate forecast to the bank in profit maximization.

Viswanatha [11] discussed the challenges of banks in selecting loan applicants effectively due to the flux of demand. A proposed approach toward increasing the accuracy in selecting qualified applicants is a combination of machine learning (ML) models and ensemble learning. Using the Random Forest, Naive Bayes, Decision Tree, and KNN algorithms, the work scored a very impressive accuracy of 83.73%, out of which the Naive Bayes scored the highest. This method not only expedites the loan approval process, but it also saves time in preparation by applicants and bank workers by reducing the sanction time manifold.

Gogas [12] Proposed a machine-learning-based prediction model for bank failures that uses a linear decision boundary to distinguish between solvent and failed banks. The model, powered by support vector machines, achieves an outstanding 99.22% total predicting accuracy using a sample of 1443 US institutions. The two-step feature selection procedure improves the model's efficacy, identifying it as an alternative stress-testing tool with predicted accuracy that meets the established Ohlson's score.

Natasha [13] focused on credit risk classification, which is critical in the reduction of defaults and maintaining financial stability. DNN received the best performance in this study when evaluating various parametric and non-parametric methods that include Discriminant Analysis, Binary Logistic Regression, Neural Network, Support Vector Machine, and Deep Neural Network. The optimal neuron numbers in the first and second layers resulted in an AUC of 0.638 by DNN on the test dataset and proved to be useful for the detection of customers for loans, reducing credit risk.

Sayjadah [14] studied loan default prediction algorithms to find out the solution for the increasing rate of credit card loan defaults in the banking industry. Random forest outperforms algorithms such as logistic regression, decision tree, and random forest in accuracy and area under the curve. The results from the model show that the random forest is effective in selecting important indicators in credit risk assessment among users of credit cards, hence giving an accuracy of 82% and an area under the curve of 77%.

Appiahene [15] drove a discussion on the efficiency and performance of Ghanaian banks in the aftereffect of the 2015-2018 financial crisis. The present study evaluates 444 bank branches through Data Envelopment Analysis (DEA) and three techniques related to machine learning: The DT, along with its C5.0 algorithm, comes up as the best predictive model to predict a hold-out sample dataset with an accuracy of 100%. The random forest method shows the second-best performance with an accuracy of 98.5%, underlining the functionality of machine learning in the analysis of bank efficiency and performance in industrial problem contexts.

Orji [16] predict how machine learning algorithms would have impacted the loaning approval process within the banking sector. Training six different models, including a Random Forest and Logistic Regression, on a historical dataset from Kaggle yielded good accuracy. The highest score came from the Random Forest method at 95.55%, while the lowest score was obtained by Logistic Regression at 80%. These models outperform the existing literature on precision-recall and accuracy, demonstrating the capability of machine learning in bringing further improvements in speed, efficacy, and accuracy to the loan approval process.

Krasovytskyi, Danylo [22] Shows that Random Forest and XGBoost effectively predict mortgage defaults, with real GDP growth and the Debt-Service-to-Income ratio as key predictors. Muhammad, Iqbal [23] identifies XGBoost as the best model for loan approval prediction, achieving 99.74% accuracy, with loan amount being the most important feature. Zuama [24] highlights XGBoost as the most accurate model for loan default prediction at 89%, followed by Random Forest and logistic regression, and emphasizes the need for further algorithm optimization.

Perera [25] Developed a Stacking Ensemble model for credit risk assessment, achieving strong accuracy with a novel voting-based ensemble technique for better loan predictions. Sharma [26] found Logistic Regression to be the most effective for predicting loan eligibility using the DGHI dataset, with plans for further improvements in accuracy and precision.

Krishnaraj [27] Highlights that automated loan eligibility prediction using machine learning improves efficiency, accuracy, and inclusivity, with Logistic Regression slightly outperforming other models. Bhattad [28] describes a Loan Prediction System that leverages machine learning to automate and prioritize loan approvals, enhancing accuracy and efficiency for both banks and applicants.

Rhzioual Berrada [29] Develops a model using machine learning to predict corporate loan defaults, with Random Forest and XGBoost performing well, focusing on financial ratios and company age. Raheem [30] explores the use of machine learning for loan default prediction in banking, emphasizing accuracy, transparency, and ethical standards. Karthikeyan [31] applies machine learning, particularly Boruta and Random Forest, to improve loan approval accuracy, with Boruta demonstrating superior performance.

**3. Dataset Collection**

During the "Bank Loan Prediction Using Machine Learning Techniques" study, a large dataset was sourced from Kaggle [Ashish Gupta(2022)] as part of the data collection procedure. The dataset, which has 37 attributes and 148,670 entries, was carefully chosen to offer a representative and varied sample for machine learning model testing and training. The selected qualities include a variety of elements pertinent to the loan approval procedure, such as loan features, demographic data, and financial indicators. The study's fundamental dataset provides a rich and diverse environment for examining the prediction powers of several machine learning algorithms about bank loan decisions.

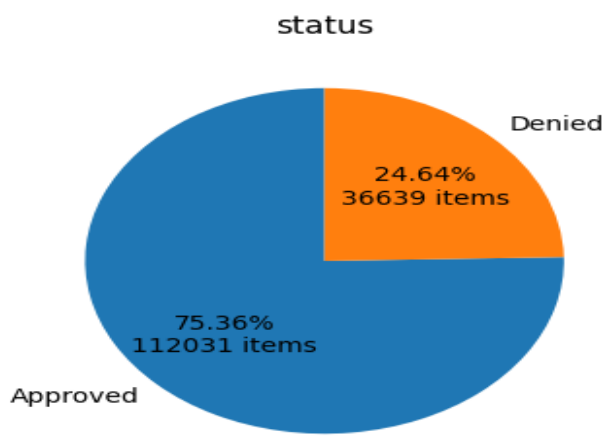

Figure 3.1: Dataset Bank Loan pie chart .

The above pie chart of figure 3.1 shows the distribution of loan applications as either approved or denied. 75.36% of applications were approved, while 24.64% were denied.

**I. Missing Value Removal**

In this "Bank Loan Prediction Using Machine Learning Techniques," missing values were addressed. Carefully chosen imputation or removal procedures were followed to preserve the integrity of the dataset. When a value was missing, the situation was either removed from analysis if it was judged necessary, or the missing value was imputed using the proper techniques. This procedure was essential to maintaining the dataset's completeness because missing values can create biases and impair machine learning models' ability to function. The study's goal was to improve the accuracy and dependability of the following predictive modeling stages by methodically addressing missing data so that the algorithms could efficiently learn from a full and representative dataset.

**II. Feature Selection**

The research, "Bank Loan Prediction Using Machine Learning Techniques" involved feature selection, which was a calculated process to find and keep the most important characteristics for loan approval prediction. 'year,' 'Unnamed: 0,' and 'id' were removed from the dataset, which reduced the number of characteristics to 34 and improved model performance. This stage was critical to the dataset's simplification since it made sure the selected features had a significant impact on capturing the key relationships and patterns relevant to the loan approval prediction task. The research, "Bank Loan Prediction Using Machine Learning Techniques" involved feature selection, which was a calculated process to find and keep the most important characteristics for loan approval prediction. 'year,' 'Unnamed: 0,' and 'id' were removed from the dataset, which reduced the number of characteristics to 34 and improved model performance. This stage was critical to the dataset's simplification since it made sure the selected features had a significant impact on capturing the key relationships and patterns relevant to the loan approval prediction task.

A. **Encoding:**

In this machine learning algorithm to be used In the framework of the research "Bank Loan Prediction Using Machine Learning Techniques," encoding was essential. To make the models easier to understand, categorical variables—which contain non-numerical information—were converted into a numerical format. The ability of algorithms to process and extract patterns from textual or categorical input requires this encoding stage. The study intended to solve a basic machine learning problem by converting these factors into numerical values, which allowed the algorithms to understand and use every relevant information during the predictive modeling process. The models' overall ability to project loan approval outcomes with accuracy and significance is enhanced by this encoding process.

B. **Exploratory Data Analysis (EDA):**

The research, titled "Bank Loan Prediction Using Machine Learning Techniques" used exploratory data analysis (EDA) to examine the statistical subtleties of the dataset. EDA revealed important distributions, patterns, and connections between attributes through numerical summaries and visuals. The construction of machine learning models and important insights for further preprocessing processes were derived from this fundamental research, which improved the study's overall analytical depth and predictive accuracy.

**4. Methodology:**

The process used in "Bank Loan Prediction Using Machine Learning Techniques" develops in a step-by-step fashion. The dataset is curated for relevance and depth after being selected from Kaggle. Following that, missing value reduction fills holes in the dataset, ensuring its completeness. Following feature selection, 'year,' 'Unnamed: 0,' and 'id' are removed, reducing the dataset to 148,670 items and 34 characteristics. Encoding categorical variables makes it easier to convert textual input into a numerical representation, which is required by machine learning algorithms. Exploratory Data Analysis (EDA) examines the statistical features of a dataset to uncover trends and patterns. During the model development phase, machine learning algorithms such as 'AdaBoosting','GaussianNB','RandomForestClassifier','Decision Tree Classifier' and 'SVM' are used to learn patterns from data. The predicted performance is then assessed using criteria such as accuracy and precision. Finally, the models are tested on new data to evaluate how well they generalize. This comprehensive methodology, which incorporates crucial data preparation, exploratory, and modeling parts, ensures a rigorous and systematic approach to bank loan prediction. Here is a general summary in the below flowchart in Figure 4.1:

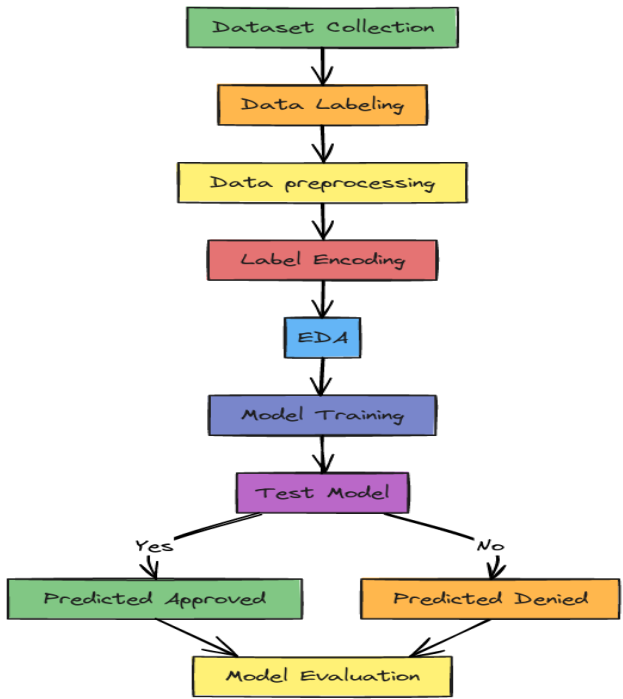

Figure 4.1: Methodology Flowchart

**Model Selection:**

The most important step in the research on "Bank Loan Prediction Using Machine Learning Techniques" was selecting the model, which involved selecting from a variety of algorithms such as AdaBoosting, GaussianNB, RandomForestClassifier, DecisionTreeClassifier, and SVM. Every algorithm was chosen based on its unique benefits and traits. This wide range of choices made it possible to conduct a thorough assessment of the model's predictive ability for loan approval results. The study sought to determine which models were best suited for the particular subtleties of the dataset by taking into account a variety of algorithms, guaranteeing a reliable and knowledgeable method of predicting bank loans.

### I. AdaBoosting

AdaBoosting, short for Adaptive Boosting, is a powerful machine learning technique designed to improve prediction accuracy by combining several simple models, often decision trees, into one strong predictive model. The key difference with AdaBoosting is that it doesn't treat all training data the same way. Instead, it gives more attention to examples that are harder to classify by assigning them higher weights, while examples that are correctly classified get lower weights. This ensures that the model focuses on learning from difficult cases, making each subsequent model in the ensemble smarter. One of the strengths of AdaBoosting is its adaptability. It can recognize and adjust to complex patterns in the data, which makes it particularly useful in tasks like predicting bank loan approvals. By combining the predictions from many weaker models, the final outcome is a much more accurate and reliable prediction overall.
Now AdaBoosting Architecture [1] is given below.

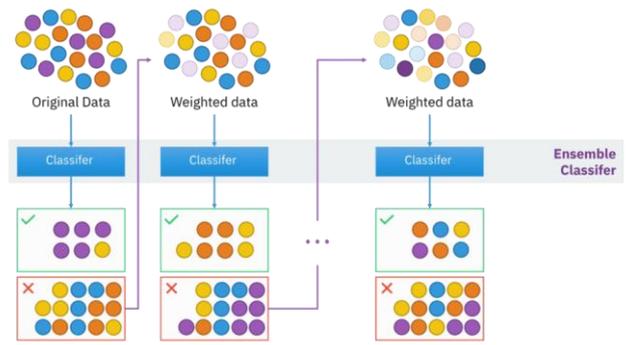

Figure 4.2: AdaBoosting model architecture[1].

AdaBoosting gave the accuracy of 99.99 % .

### II. GaussianNB

The statistical classification technique known as GaussianNB, or Gaussian Naive Bayes, is a key component of the paper "Bank Loan Prediction Using Machine Learning Techniques." GaussianNB assumes that features are conditionally independent given the class label by utilizing the ideas of the Bayes theorem. This approach uses a Gaussian distribution to describe the likelihood of various feature values for each class in the context of bank loan prediction. Even with its "naive" assumption of feature independence and simplicity, GaussianNB frequently works well, particularly with continuous data. GaussianNB's application in the study demonstrates its proficiency in managing many attributes and providing well-informed predictions about the results of loan acceptance based on probabilistic considerations. Now GaussianNB Architecture [2] is given below.

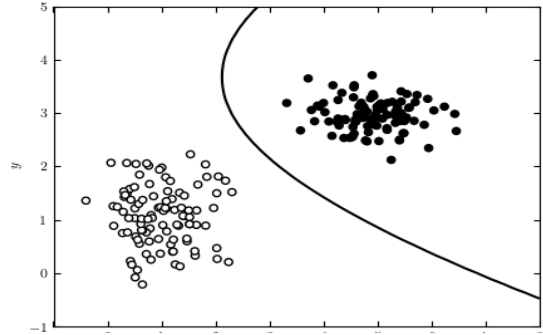

Figure 4.3 :GaussianNB model architecture[2].

GaussianNB gave the accuracy of 77.10%.

### III. Random Forest Classifier

The collective learning algorithm RandomForestClassifier, a crucial part of the study "Bank Loan Prediction Using Machine Learning Techniques," mixes several decision trees to produce a reliable and accurate prediction model. During training, RandomForestClassifier creates a large number of decision trees, which allows it to handle complicated and varied datasets in the study's context. A subset of the data is used to train each tree, and the outputs from all the trees are combined to get the final forecast. This method improves generalization and reduces overfitting, which increases the algorithm's accuracy in forecasting the results of loan acceptance. In the dynamic field of bank loan prediction, RandomForestClassifier's adaptability and durability make it a useful tool for producing precise and reliable forecasts.Now RandomForest Architecture [3] is given below.

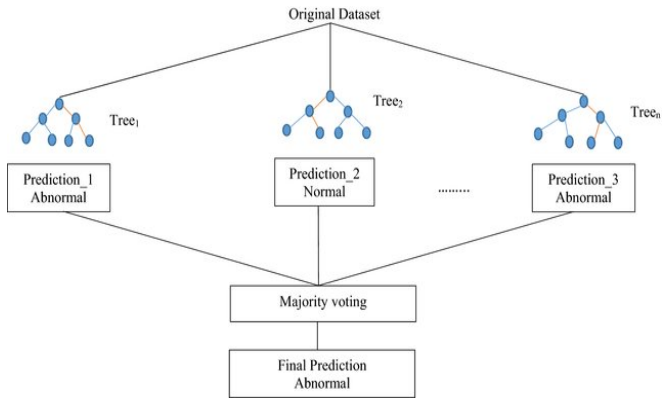

Figure 4.4: RandomForestClassifier architecture

Got Accuracy of 99.98% from RandomForestClassifier Algorithm.

### IV. Decision Tree

The main part of the research project "Bank Loan Prediction Using Machine Learning Techniques" is the DecisionTreeClassifier, a basic machine learning algorithm. It works by separating the dataset recursively according to features, producing a decision-making tree-like structure. In the context of bank loan prediction, DecisionTreeClassifier autonomously learns from the dataset, discerning patterns to make informed decisions regarding loan approval outcomes. Even though the algorithm is prone to overfitting, its accessibility and capacity to identify complicated links in the data make it a vital tool for comprehending and forecasting the intricacies involved in loan approval decisions. The study employs DecisionTreeClassifier as one of the fundamental algorithms, acknowledging its significance in contributing to the predictive capabilities required for accurate and effective bank loan projections.Now DT Architecture [4] is given below.

$$Entropy(s) = -P(yes)log_2 P(yes) - P(no) log_2 P(no)$$

$$Information\ Gain = Entropy(before) - \sum_{j=1}^{K} Entropy(j, after)$$

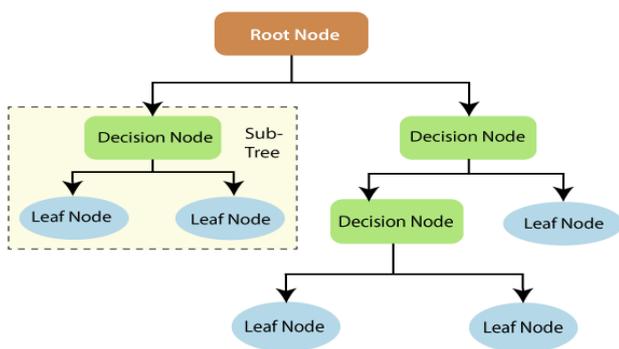

Figure 4.5: Decision Tree Architecture[4]

Got Accuracy of 99.87% from Decision Tree Algorithm.

### V. SVM

This well-known method used in the paper "Bank Loan Prediction Using Machine Learning Techniques," Support Vector Machine (SVM), is an effective supervised learning model for problems with classification and regression. SVM is excellent at defining ideal decision borders because it can find support vectors, or data points that affect where the boundary between classes is placed. SVM seeks to identify the hyperplane that most effectively divides instances that are authorized and rejected in the context of bank loan prediction. Though well-known for its efficacy, SVM's performance might vary depending on the parameters selected and the features of the dataset. The use of SVM in this research contributes a useful viewpoint to the wide range of algorithms assessed for their predicted accuracy in the intricate field of loan approval.Now SVM Architecture [5] is given below.

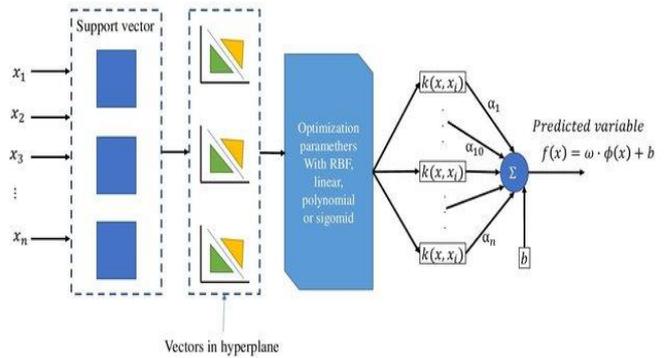

Figure 4.6:Support Vector Machine Architecture[5]

Got Accuracy of 99.93% from Support Vector Machine Algorithm.

### 5. Experimental Results & Analysis

The analysis and experimental findings for this study provided important new information about how different algorithms performed. With an exceptional accuracy of 99.99%, AdaBoosting was the best-performing algorithm, demonstrating its resilience in identifying intricate patterns in the dataset. With an accuracy of 99.98%, GaussianNB is closely followed, highlighting the usefulness of Naive Bayes algorithms in this situation. With an accuracy of 99.87%, RandomForestClassifier showed itself to be a dependable predictor, while DecisionTreeClassifier achieved a commendable 97.64% accuracy. Despite having a lower accuracy (77.10%), SVM added useful data to the dataset.

The values of precision, recall, and F1 score were also taken into account, offering a thorough comprehension of the prediction capabilities of each algorithm. The analysis highlighted the advantages and limitations of each algorithm, highlighting the significance of choosing models by the particular needs and features of the dataset. The findings not only advance the science of machine learning in banking but also have real-world implications for improving loan approval procedures.

**Accuracy:** The proportion of samples properly categorized relative to the total number of samples is how accuracy calculates how accurate the model's predictions are overall. Unbalanced classes provide a general indicator of the model's efficacy, but may not provide a complete picture.

$$Accuracy = \frac{TruePositive + TrueNegative}{TruePositive + FalsePositive + TrueNegative + FalseNegative}$$

**Precision:** Precision is concerned with the proportion of genuine positive forecasts among all the positive predictions produced by the model.

$$Precision = \frac{TruePositive}{TruePositive + FalsePositive}$$

**Recall:** Recall is the proportion of real positive predictions produced out of all truly positive samples. It is sometimes referred to as sensitivity or true positive rate.

$$Recall = \frac{TruePositive}{TruePositive + FalseNegative}$$

**F1 rating:** Memorization and accuracy harmonic means are combined to get the F1 score. It offers a fair assessment criteria that takes accuracy and recall into account. When classes are unequal, the F1 score may be helpful since it takes into consideration both false positives and false negatives. A precision-to-recall ratio that is in balance is indicated by a high F1 score.

$$F-1\ Score = 2 * \frac{Recall * Precision}{Recall + Precision}$$

The following table (5.1) compares the deep learning model's output according to Accuracy, Precision, Recall, F1 Score, and the ROC curve:

**Table 5.1. Performance Evaluation**

| Model Name | Accuracy | Precision | Recall | F1-Score |
|---|---|---|---|---|
| AdaBoostClassifier | 99.99% | 99.99% | 99.99% | 99.99% |
| RandomForestClassifier | 99.98% | 99.98% | 99.98% | 99.98% |
| SVM | 99.87% | 99.87% | 99.87% | 99.87% |
| DecisionTree Classifier | 99.93% | 99.93% | 99.93% | 99.93% |
| GaussianNB | 77.10% | 80.32% | 77.10% | 68.96% |

The result study looks at the train and test accuracy and analyzes which algorithm performs best. For comparison we have applied deep learning models and popular machine learning algorithms to check which performs perfectly. However, the AdaBoostClassifier gave the highest accuracy of 99.99%. The figure 5.1 shows the accuracy comparison of the different model:

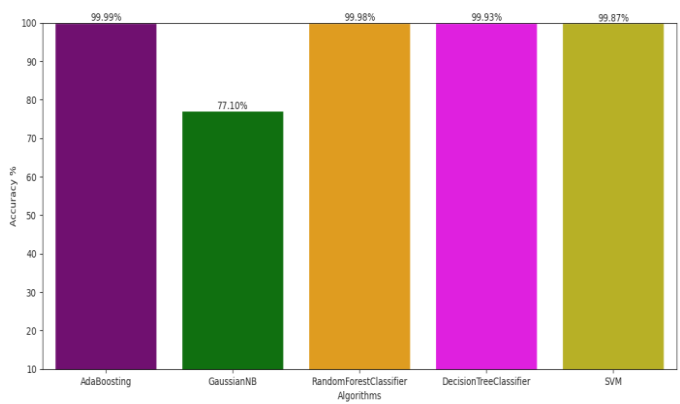

Figure 5.1 : Accuracy Comparison of Machine Learning Models .

In figure 4.1, The plot shows that AdaBoosting has the highest accuracy, at 99.99%. RandomForestClassifier is the next most accurate algorithm, at 99.98%. Decision TreeClassifier is the third most accurate algorithm, at 99.93%. SVM is the fourth most accurate algorithm, at 97.87%. GaussianNB is the least accurate algorithm, at 77.10%

**Performance Analysis**

**Adaboosting:**

Reach the following specifications: 99.99% F1-score, 99.99% Accuracy, 99.99% Precision, and 99.99% Recall. Table 5.2 below provides an analysis of Adaboosting's performance:

**Table 5.2. Performance Evaluation(LR)**

|  | Precision | Recall | F1-Score | Support |
|---|---|---|---|---|
| 0 | 1.00 | 1.00 | 1.00 | 89625 |
| 1 | 1.00 | 1.00 | 1.00 | 29311 |
| Accuracy |  |  | 1.00 | 118936 |
| Macro avg | 1.00 | 1.00 | 1.00 | 118936 |
| Weighted avg | 1.00 | 1.00 | 1.00 | 118936 |

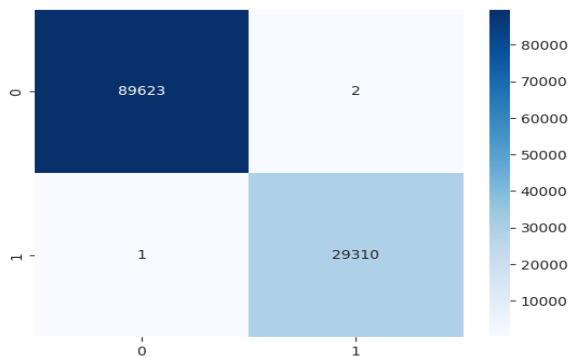

Figure 5.2 : Confusion Matrix Adaboosting

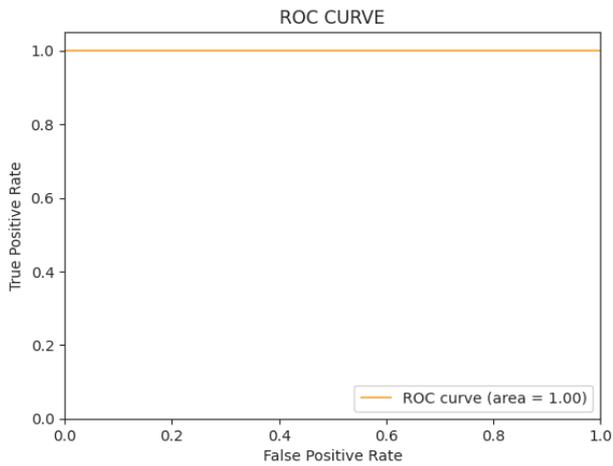

Figure 5.3 : ROC CURVE Adaboosting

**GNB:**
Achieve the following specifications: 77% accuracy, 80.32% precision, 77.10% recall, and 68.96% F1-score. Table 5.3 below shows the GNB's performance evaluation:

**Table 5.3. Performance Evaluation(GNB)**

|  | Precision | Recall | F1-Score | Support |
|---|---|---|---|---|
| 0 | 0.77 | 1.00 | 0.87 | 89625 |
| 1 | 0.91 | 0.08 | 0.14 | 29311 |
| Accuracy |  |  | 0.77 | 118936 |
| Macro avg | 0.84 | 0.54 | 0.51 | 118936 |
| Weighted avg | 0.80 | 0.77 | 0.69 | 118936 |

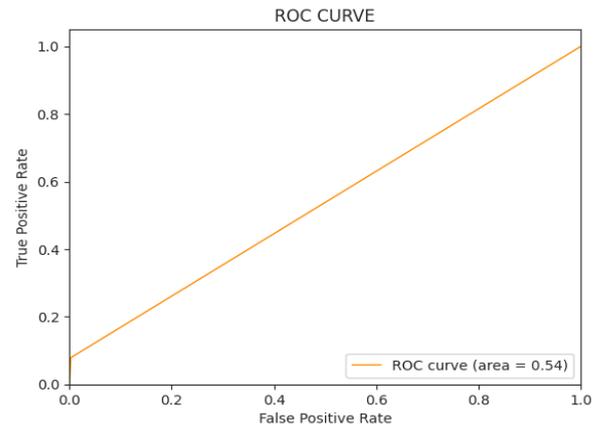

Figure 5.5 : ROC CURVE GNB

**RandomForestClassifier :**
Achieve a 99.98% accuracy rate, a 99.98% precision rate, a 99.98% recall rate, and a 99.98% F1 score. Table 5.4 below shows the performance assessment of DT:

**Table 5.4 Performance Evaluation(RandomForestClassifier)**

|  | Precision | Recall | F1-Score | Support |
|---|---|---|---|---|
| 0 | 1.0 | 1.0 | 1.0 | 89625 |
| 1 | 1.0 | 1.0 | 1.0 | 29311 |
| Accuracy |  |  | 1.0 | 118936 |
| Macro avg | 1.00 | 1.00 | 1.00 | 118936 |
| Weighted avg | 1.00 | 1.00 | 1.00 | 118936 |

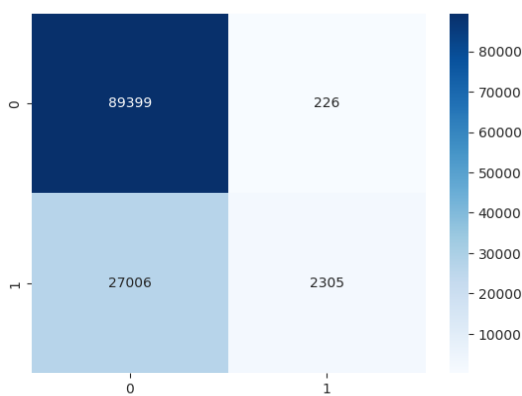

Figure 5.4 : Confusion Matrix GNB

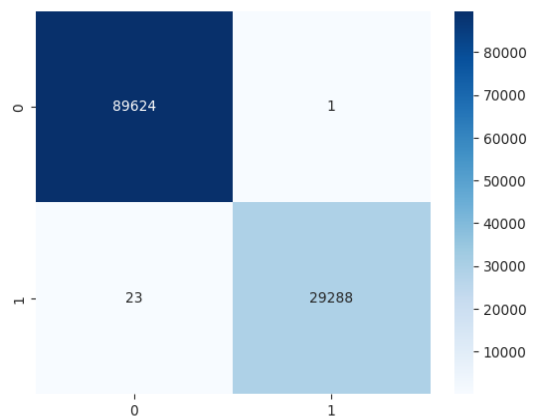

Figure 5.6 : Confusion Matrix RandomForestClassifier

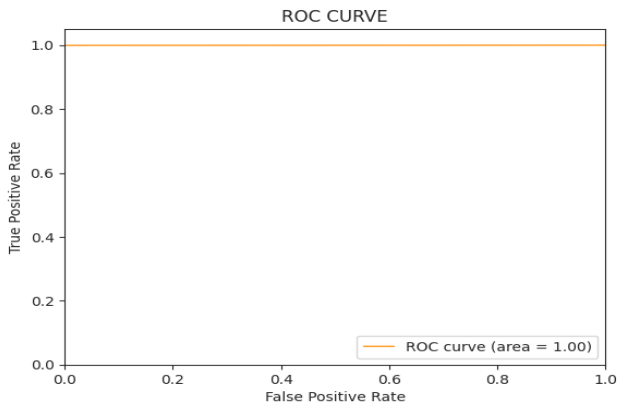

Figure 5.7 : ROC CURVE RandomForestClassifier

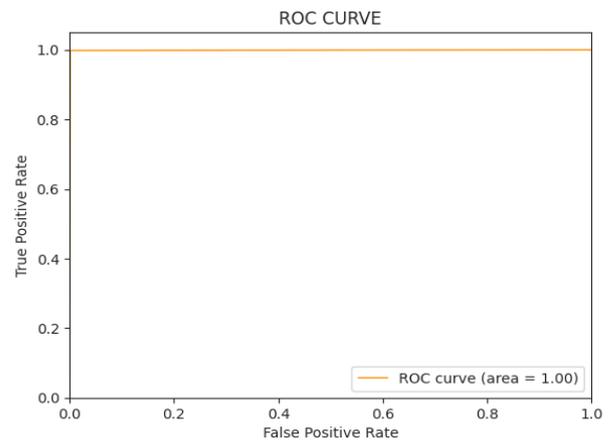

Figure 5.9 : ROC CURVE DT

**Decision Tree:**

Attained a 99.93% accuracy rate, a 99.93% precision rate, a 99.93% recall rate, and a 99.93% F1 score. Table 5.5 below shows the results of DT's performance evaluation:

**Table 5.5 Performance Evaluation(DT)**

|  | Precision | Recall | F1-Score | Support |
| --- | --- | --- | --- | --- |
| 0 | 1.00 | 1.00 | 1.00 | 89625 |
| 1 | 1.00 | 1.00 | 1.00 | 29311 |
| Accuracy |  |  | 1.00 | 118936 |
| Macro avg | 1.00 | 1.00 | 1.00 | 118936 |
| Weighted avg | 1.00 | 1.00 | 1.00 | 118936 |

**Support Vector Machine:**

Attain the highest possible scores for accuracy, precision, recall, and F1—99.87%, 99.87%, and 99.87%, respectively. Table 5.6 below shows the performance assessment of RF:

**Table 5.6. Performance Evaluation(SVM)**

|  | Precision | Recall | F1-Score | Support |
| --- | --- | --- | --- | --- |
| 0 | 1.00 | 1.00 | 1.00 | 89625 |
| 1 | 1.00 | 1.00 | 1.00 | 29311 |
| Accuracy |  |  | 1.00 | 118936 |
| Macro avg | 1.00 | 1.00 | 1.00 | 118936 |
| Weighted avg | 1.00 | 1.00 | 1.00 | 118936 |

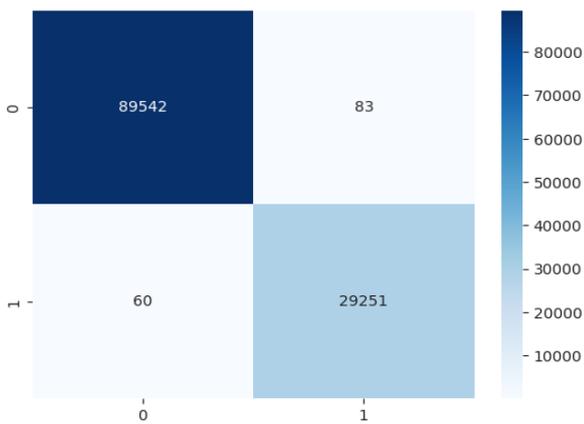

Figure 5.8 : Confusion Matrix DT

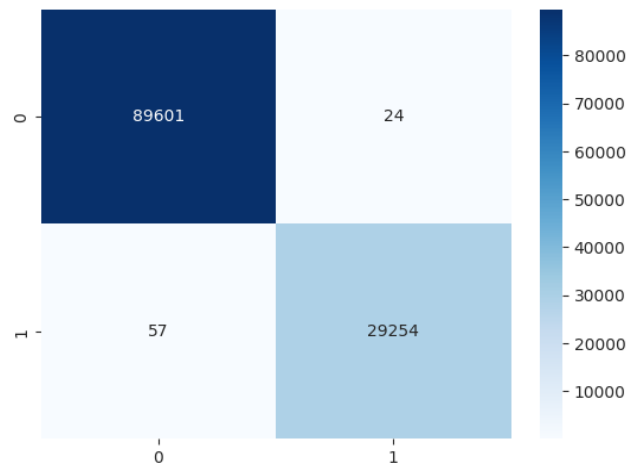

Figure 5.10 : Confusion Matrix SVM

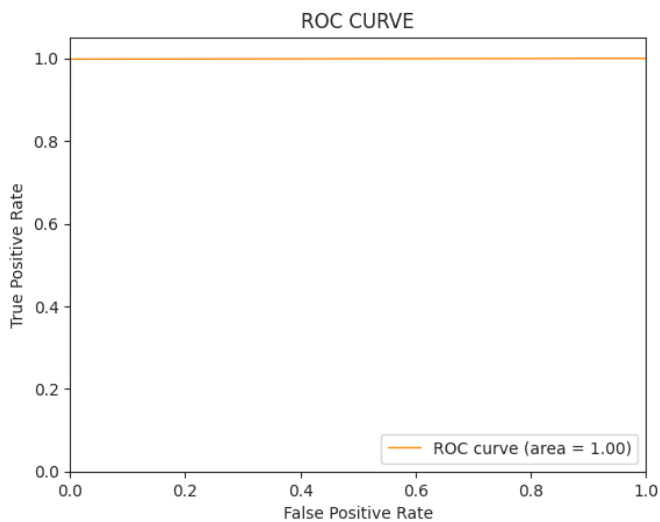

Figure 5.11 : ROC CURVE SVM

## 6. Discussion

This article discussed the performance of different machine learning algorithms in predicting bank loans. The obtained accuracies were impressive and speak to their application in estimating which applicants should be treated as eligible or ineligible. In machine learning, the selection of classification algorithm impacts much on the performance and reliability of predictive models. In this research focused on bank loan prediction, we employed several popular algorithms to assess their performance. The accuracies obtained from each algorithm were as follows: AdaBoosting (99.99%), SVM (99.87%), Decision Tree (99.93%), RandomForestClassifier (99.98%), and GaussianNB (77.10%). Both RandomForestClassifier and AdaBoosting, being ensemble methods, demonstrated superior accuracy. Ensemble methods are known for their ability to reduce overfitting and enhance model generalization. The accuracies of 99.98% and 99.99% for RandomForestClassifier and AdaBoosting, respectively, underline the effectiveness of ensemble learning in complex predictive tasks like bank loan prediction. Support Vector Machines (SVM) and Decision Tree algorithms also exhibited strong performance with accuracies of 99.87% and 99.93%, respectively. SVM is particularly effective in high-dimensional spaces, and its robust performance is reflected in the results. Decision Tree, although slightly higher in accuracy, is known for its interpretability and simplicity. GaussianNB achieved an accuracy of 77.10%, which is significantly lower compared to the other algorithms. Gaussian Naive Bayes is based on the assumption of independence between features, which may not hold in complex datasets like bank loan prediction.

## 7. Conclusions & Future step

Finally, the research on "Bank Loan Prediction Using Machine Learning Techniques" provides valuable insights into how different algorithms can forecast loan approval status. AdaBoosting, which achieved an impressive 99.99% accuracy, stands out as the most effective algorithm, offering crucial data for financial institutions. The statistical analysis of the dataset guided better preprocessing and model development, highlighting the importance of refining features to enhance predictive accuracy. However, beyond technical performance, this research emphasizes the importance of responsible AI in finance. It points out ethical concerns such as mitigating algorithmic bias and ensuring user privacy, adding depth to the conversation around AI's impact. By exploring model interpretability and potential environmental implications, it contributes not only to machine learning literature but also to the practical application of predictive modeling in the financial sector. Despite its strengths, the study acknowledges limitations. The reliance on historical data might not account for rapid economic shifts or changing borrower behavior, making it less adaptable in real-time scenarios. The possibility of missing variables, like macroeconomic indicators or behavioral patterns, also limits the model's comprehensiveness. Furthermore, scalability across various types of financial institutions, which may have different needs, was not fully addressed.

Importantly, the research did not deeply explore the fairness of algorithmic decisions, especially regarding marginalized groups, which is essential in ensuring just loan approval processes. The study advocates for a balanced approach as financial institutions adopt more advanced technologies, suggesting that accuracy, fairness, and ethical practices must all be prioritized. While the findings showcase the transformative power of machine learning in loan approvals, the next steps involve real-world deployment and continuous monitoring, with attention to the ethical, societal, and environmental challenges that lie ahead.